\pdfoutput=1
\documentclass[letterpaper, 10 pt, journal, twoside]{IEEEtran}
\IEEEoverridecommandlockouts
\usepackage{amsmath,amssymb, amsfonts, bm} 
\usepackage{url}
\usepackage{doi}
\usepackage{newclude}
\usepackage{graphicx}
\usepackage{tabularx}
\usepackage[font={small}]{caption}
\usepackage{subcaption}
\usepackage{dblfloatfix}
\usepackage[linesnumbered,ruled,vlined]{algorithm2e}
\usepackage{multirow}
\usepackage{color}
\usepackage{cite}
\usepackage{hyperref}   

\newcommand{\trsp}{{\scriptscriptstyle\top}}

\begin{document}
\title{Logic Learning from Demonstrations for Multi-step Manipulation Tasks in Dynamic Environments}
\author{Yan Zhang, Teng Xue$^{*}$, Amirreza Razmjoo$^{*}$, Sylvain Calinon%
\thanks{This work was supported by the State Secretariat for Education, Research and Innovation in Switzerland for participation in the European Commission’s Horizon Europe Program through the INTELLIMAN project (\url{https://intelliman-project.eu/}, HORIZON-CL4-Digital-Emerging Grant 101070136) and the SESTOSENSO project (\url{http://sestosenso.eu/}, HORIZON-CL4-Digital-Emerging Grant 101070310). We also acknowledge support from the China Scholarship Council (Grant No. 202106230104) and the SWITCH project  (\url{https://switch-project.github.io/}), funded by the Swiss National Science Foundation.}%

\thanks{The authors are with the Idiap Research Institute, CH-1920 Martigny, Switzerland and also with the EPFL, 1015 Lausanne, Switzerland; Authors with * contributed equally. E-mail:firstname.lastname@idiap.ch}%

}    


\maketitle

\begin{abstract}
    Learning from Demonstration (LfD) stands as an efficient framework for imparting human-like skills to robots. Nevertheless, designing an LfD framework capable of seamlessly imitating, generalizing, and reacting to disturbances for long-horizon manipulation tasks in dynamic environments remains a challenge. To tackle this challenge, we present Logic-LfD, which combines Task and Motion Planning (TAMP) with an optimal control formulation of  Dynamic Movement Primitives (DMP), allowing us to incorporate motion-level via-point specifications and to handle task-level variations or disturbances in dynamic environments. We conduct a comparative analysis of our proposed approach against several baselines, evaluating its generalization ability and reactivity across three long-horizon manipulation tasks. Our experiment demonstrates the fast generalization and reactivity of Logic-LfD for handling task-level variants and disturbances in long-horizon manipulation tasks.
    
    Project webpage: \href{https://sites.google.com/view/logic-dmp/%E9%A6%96%E9%A1%B5}{\textcolor{blue}{https://sites.google.com/view/logic-lfd}}
\end{abstract}


\renewcommand{\textcolor}[2]{#2}

\IEEEpeerreviewmaketitle

\section{Introduction}
\IEEEPARstart{L}{earning} from Demonstrations (LfD) has proven to be an effective approach for enabling robots to tackle complex manipulation tasks by imitating expert demonstrations \cite{Billard16chapter}.
Recent advancements in LfD have extended its applicability to long-horizon manipulation tasks, such as table rearrangement or kitchen activities, by segmenting demonstrations into sub-tasks \cite{mandlekar2020learning}, keyframes \cite{perez2017c}, or skills \cite{konidaris2012robot}. However, existing works often focus solely on reproducing demonstrations, assuming a sequential execution of learned skills can achieve task goals, which is not always applicable in real-world scenarios \cite{Wang_Figueroa_Li_Shah_Shah_2022}. While traditional LfD methods like Dynamic Movement Primitives (DMP) \cite{ijspeert2013dynamical}, Task-parameterized Gaussian Mixture Model (TP-GMM) \cite{calinon2016tutorial}, or Dynamical Systems (DS) \cite{billard2022learning} can generalize demonstrated trajectories for new tasks or react to disturbances at the motion level, long-horizon planning necessitates not only motion-level generalization but also the ability to handle task-level variations and disturbances. Consequently, designing an LfD framework that can imitate, generalize, and react to disturbances while solving multi-step manipulation tasks remains a challenge \cite{mandlekar2020learning, Wang_Figueroa_Li_Shah_Shah_2022}.

In the domain of long-horizon planning, Task and Motion Planning (TAMP) has emerged as a powerful tool for solving multi-step manipulation tasks \cite{garrett2020pddlstream, toussaint2015logic}. TAMP involves searching over a set of abstracted actions, with parameters often determined through sampling- or gradient-based motion planning. While theoretically capable of solving any feasible long-horizon manipulation task, TAMP relies on accurate dynamics modeling \cite{mandlekar2023human} and combinatorial searching of task instances and valid motions \cite{mandlekar2020learning}. This limits the application of TAMP methods in real-world manipulation scenarios, involving model uncertainty and external disturbances.

\begin{figure}[t]
    \centering
    \includegraphics[trim=0.0cm 5.1cm 0.0cm 5.1cm, clip, width=\linewidth]{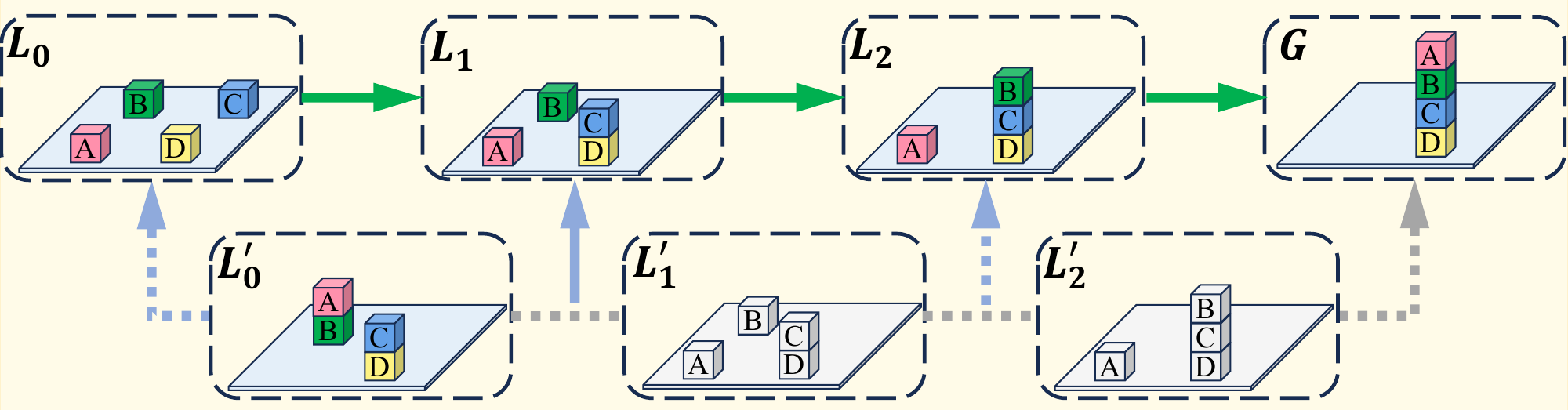}
    \caption{Overview of Logic-LfD. Arrows refer to action primitives encoded with the proposed DMP variant (LQT-CP). The template task starts from $\bm{\mathcal{L}_{0}}$ and ends at goal $\bm{G}$. $\bm{\mathcal{L}_{0}} \rightarrow \bm{\mathcal{L}_{1}} \rightarrow  \bm{\mathcal{L}_{2}} \rightarrow \bm{G}$ illustrates the task-level demonstration. For a new task starting from $\bm{\mathcal{L}_{0}^{\prime}}$, a \textcolor{blue}{fixed sequential} execution of actions primitives  encoded by DMPs (green arrows) cannot transition from $\bm{\mathcal{L}_{0}^{\prime}}$ to the goal state $\bm{G}$. Typical TAMP solvers find the action skeleton from $\bm{\mathcal{L}_{0}^{\prime}}$ to $\bm{G}$ from scratch (grey dashed long arrow). Instead, Logic-LfD tries to reach \textcolor{blue}{both task goal $\bm{G}$ and all other states (blue arrows)} in the demonstration in parallel \textcolor{blue}{to find} a feasible plan $\bm{\mathcal{L}_{0}^{\prime}}$ $\rightarrow$ $\bm{\mathcal{L}_{1}}$ connecting $\bm{\mathcal{L}_{0}^{\prime}}$ to the demonstration trajectory within the \textcolor{blue}{minimum} time. It then merges the new plan with the corresponding segmentation of the demonstration $\bm{\mathcal{L}_{1}} \rightarrow  \bm{\mathcal{L}_{2}} \rightarrow \bm{G}$, thus accomplishing the new task faster than classical TAMP solvers.
    }
    \label{fig:framework}
\end{figure}

In this work, we propose Logic-LfD, an LfD approach that generalizes one multi-step demonstration to solve new similar multi-step tasks, by \textcolor{blue}{combining} the advantages of DMP and TAMP solvers. Logic-LfD leverages an optimal control formulation of DMP for motion modulation and extends it to incorporate via-point specifications for solving contact-rich manipulation sub-tasks, like pulling a cube with a hook in Figure.~\ref{fig:lqt_vs_dmp}.
\textcolor{blue}{Combining} LfD with TAMP solvers can mitigate the challenges associated with modeling complex contact dynamics. Besides, we show that the demonstration can be formulated into multi-goal specifications for \textcolor{blue}{the} TAMP solver so that it can try to achieve all the states on the demonstrated task-and-motion trajectory in parallel, which significantly accelerates the planning process while generalizing the demonstration for solving new tasks. We further extend our approach to introduce a reactive TAMP framework by deploying Logic-LfD in a closed-loop fashion for real-world tasks in dynamic environments.

Notably, our choice of DMP as the motion modulation block is motivated by its extrapolation generalization ability based on a single demonstration. However, conventional DMP suffers from poor \textcolor{blue}{via-point} encoding ability, limiting its applicability for tasks with specific \textcolor{blue}{via-point} requirements that can facilitate collision avoidance or contact-rich manipulation tasks \cite{zhou2019learning, mghames2020interactive}. In our prior work \cite{Calinon_IET_2023}, we propose Linear Quadratic Tracking with Control Primitives (LQT-CP), an optimal control formulation of DMP that can freely modulate \textcolor{blue}{how} the system reacts to perturbations. In this paper, we explore the flexible motion modulation abilities of LQT-CP and further extend it to incorporate \textcolor{blue}{via-point} tracking and generalization capability, addressing the \textcolor{blue}{limitations} inherent in classical DMP.

In summary, our work proposes the following technical advancements:
\begin{itemize}
    \item \textcolor{blue}{We propose Logic-LfD, a novel integration of TAMP solver with classical LfD methods. This integration extends DMP to solve long-horizon manipulation tasks by imitating multi-step demonstrations. It also explores the capability of LQT-CP to extend DMP to incorporate via-point tracking and generalization for addressing contact-rich manipulation sub-tasks;
    \item Logic-LfD improves the generalization ability of DMP to handle both motion- and task-level variants while solving similar multi-step manipulation tasks;}
    \item \textcolor{blue}{We develop a Reactive TAMP approach, leveraging the fast generalization capability of Logic-LfD. This approach improves the reactivity of DMP to task-level disturbances, thus facilitating the solution of long-horizon manipulation tasks in dynamic environments.}
\end{itemize}

\section{Related Work}

\subsection{LfD for long-horizon manipulation tasks}
Previous LfD methods typically address long-horizon manipulation tasks by segmenting demonstrations and composing movement primitives trained with the segmented demonstration \cite{saveriano2019merging, niekum2013incremental}. \textit{Schwenkel et al.} and \textit{Jaquier et al.} propose methods to find the best sequence of primitives based on demonstrations for the target tasks \cite{Schwenkel_Guo, jaquier2022learning}. Our Logic-LfD shares similarities in composing skill primitives for complex long-horizon manipulation tasks. However, unlike those methods, it does not assume a \textcolor{blue}{fixed} sequential composition of those primitives (namely, without generalization or replanning) \textcolor{blue}{while solving the target long-horizon manipulation task in the real world}. We show in Experiments \textcolor{blue}{(Section. \ref{sec:experiments}) with Tables \ref{tab:generalization} and \ref{tab:reactive}} that the removal of this assumption can improve the generalization ability and reactivity of the system to handle task variants and disturbances\textcolor{blue}{, facilitating long-horizon manipulation planning in dynamic environments.}

Under the umbrella of \textcolor{blue}{combining} LfD with TAMP, \textit{Mandlekar et al.} propose human-in-the-loop TAMP where human teleoperation assists TAMP solvers \textcolor{blue}{in finding} feasible motions for complex contact-rich manipulation tasks without complicated dynamics modeling \cite{mandlekar2023human}. \textit{Mcdonald et al.} and \textit{Dalal et al.}\cite{mcdonald2022guided, Dalal_Mandlekar_Garrett_Handa_Salakhutdinov_Fox_2023} propose to train a policy imitating a TAMP planner with a large-scale database of demonstrations collected offline, achieving good performance on generalization and reactivity. \textcolor{blue}{Moreover, large language models (LLMs) have proven beneficial for high-level task planning for long-horizon manipulation tasks \cite{firoozi2023foundation, ahn2022can}. They can be considered as human language behavior models learned from internet-scale demonstrations. However, these language behavior models lack real-world experiences and robot information \cite{Xiang2023Language}, necessitating great efforts on domain data collection and fine-tuning to find feasible trajectories for solving multi-step manipulation tasks with TAMP solvers in the real world \cite{driess23palme}. In contrast, our Logic-LfD can handle various long-horizon manipulation tasks by imitating and generalizing a single demonstration.}

\subsection{Reactive Task and Motion Planning}
Our closed-loop Logic-LfD operates within the domain of reactive TAMP, where rapid replanning is essential to respond to disturbances in dynamic environments. In recent studies \cite{Wang_Figueroa_Li_Shah_Shah_2022} \cite{Li_Park_Sung_Shah_Roy_2021}, temporal logic-based reactive synthesis has been combined with Dynamical Systems (DS) or behavior tree-based control strategies for reactive action selection and plan switching to address task-level disturbances in multi-step manipulation tasks. In contrast, our approach delves into a PDDL-based (Planning Domain Definition Language) TAMP planning framework, which includes logical state and action abstraction, thus facilitates the integration of action primitives learned from demonstration. 

In \cite{Migimatsu_Bohg_2020}, \textit{Migimatsu et al.} propose to execute motions of the TAMP plan in object-centric Cartesian coordinates, demonstrating efficiency in reacting to motion-level disturbances. However, their method does not extend to handling task-level disturbances, setting it apart from our approach. In \cite{xue2023d}, \textit{Xue et al.} introduce the Dynamic Logic-Geometric Program (D-LGP), showcasing impressive time efficiency for TAMP. However, D-LGP relies on keyframe-based action skeleton planning, making it incapable of handling disturbances within the keyframes. Receding-Horizon TAMP (RH-TAMP) approaches iteratively solve a reduced planning problem over a receding window of \textcolor{blue}{a} shorter action skeleton, accelerating the TAMP planning process for handling interventions in changing environments \cite{castaman2021receding, Braun_Ortiz_Haro_Toussaint_Oguz_2022}. Nevertheless, a shorter prediction horizon increases the undesired infeasibility of action skeletons. Our Logic-LfD also accelerates TAMP by reducing the planning problem into a partial problem. However, it does not adopt a receding-horizon fashion, thus avoiding the influence on the feasibility.

In \cite{Harris_Driess_Toussaint_2022}, \textit{Harris et al.} propose a Feasibility-based Control Chain Coordination (FC$^{3}$) approach, showcasing impressive reactiveness by constructing offline a set of possible action plans and by switching between them in an online manner. However, the success of FC$^{3}$ heavily relies on the constructed plan library and switching strategy. The closest work to ours is Robust Logical-Dynamical Systems (RLDS) \cite{Paxton_Ratliff_Eppner_Fox_2019}, which modifies the initial action plan by jumping backward or forward. However, RLDS assumes task disturbances can be handled without action plan switching, significantly limiting its reactiveness under substantial task perturbations. In several following experiments, our method demonstrates superior reactiveness to various levels of task disturbances than RLDS, which is achieved by \textcolor{blue}{an} online generalization of the demonstrated nominal plan, without the need of constructing an action plan library offline as in \cite{Harris_Driess_Toussaint_2022}.

\section{Preliminary}
\subsection{Planning Domain Definition Language (PDDL)}
In this section, we present essential concepts related to the Planning Domain Definition Language (PDDL) using the block stacking task illustrated in Figure~\ref{fig:framework} as an example. PDDL serves as an interface for defining the world by specifying a set of facts that describe relationships among objects in the environment. For instance, in $\mathcal{L}_{1}$ of Figure~\ref{fig:framework}, the cube C is positioned on the cube D and this relationship is represented by the fact \textit{(on C D)}. Throughout this paper, we use italicized symbols to denote these facts and the term \textit{scene graph} to collectively represent the entire set of facts.

Among these facts, those that remain constant throughout the planning process \textcolor{blue}{is} termed static facts (e.g., \textit{(cube C)}). Conversely, facts that vary during the process are referred to as fluents (e.g., \textit{(on C D)}, which changes with the robot's operation).

PDDL also provides the concept of action abstraction. An action is defined by a set of parameters, preconditions that must be satisfied before executing the action, and effects that occur after the action. Taking the action \textit{pick} as an example, its parameters may include specifying which robot(s) pick which object(s). If, for instance, we have \textit{(robot panda)} and \textit{(cube C)} as the parameters, one precondition for executing this action should be \textit{(clear C)}, indicating that there are no cubes on C, making it possible to grasp. The resultant effect is denoted by \textit{(inhand panda C)}, signifying that cube C is now in the hand of the robot arm panda.

\section{Method}

\subsection{LQT-CP: an optimal control formulation of DMP}

DMP is constructed in two parts to converge to the final goal position while tracking the acceleration profiles for mimicking the shape of a demonstrated trajectory. Similarly, Linear Quadratic Tracking (LQT), the most basic form of optimal control, is described by a cost typically composed of several parts acting at different state and control levels, with references either in the form of full trajectory or final goal positions. In particular, the cost can be specified to track a reference velocity or acceleration profile while reaching a target state at the end of the movement, with weight matrices balancing the importance of tracking the desired profiles and reaching the final goal position. With this similarity, we can reformulate DMP into a LQT fashion with the acceleration profile of the demonstrated trajectory as \textcolor{blue}{the} reference trajectory and the attractor as the goal to be reached at the end.

Similarly to DMP, our system is a virtual point mass driven by a linear integrator system to describe the evolution of the system. As shown in \cite{Calinon_IET_2023}, the control profile can be encoded with basis functions, resulting in a Control Primitive (CP) formulation of LQT that estimates superposition weights instead of the full list of control commands. The resulting LQT-CP yields a DMP by minimizing the cost function:
\begin{equation}\label{eq:batch_formulation_lqt}
    \begin{split}
        c = (\bm{\mu} - \bm{x})^{\top} \bm{Q} (\bm{\mu} - \bm{x}) + \bm{u}^{\top} \bm{R} \bm{u},
    \end{split}
\end{equation}
where $\bm{\mu}=[\bm{\mu}^\trsp_{0}, \bm{\mu}^\trsp_{1}, \ldots, \bm{\mu}^\trsp_{T}]^\trsp$ indicates the concatenated vector of the position, velocity, and acceleration profiles of the demonstrated trajectory and $\bm{\mu}_{T}$ includes the goal position $\bm{g}$ to be reached at the end. $\bm{x}=[\bm{x}^\trsp_{0}, \bm{x}^\trsp_{1}, \ldots, \bm{x}^\trsp_{T}]^\trsp$ represents the concatenated system state variables. The matrices $\bm{Q} = \text{diag}(\bm{Q}_{0}, \bm{Q}_{1}, \ldots, \bm{Q}_{T})$ and $\bm{R} = \text{diag}(\bm{R}_{0}, \bm{R}_{1}, \ldots, \bm{R}_{T})$ are a block-diagonal matrices showing the evolution of weight matrices $\bm{Q}_{t}$ and $\bm{R}_{t}$ for variable $\bm{x}$ and command $\bm{u} = [\bm{u}^\trsp_{0}, \bm{u}^\trsp_{1}, \ldots, \bm{u}^\trsp_{T}]^\trsp$, correspondingly. The evolution of the state is described by a linear integrator $\bm{x}_{t+1}=\bm{A}_t \bm{x}_t+\bm{B}_t \bm{u}_t$, yielding $\bm{x}=\bm{S}_{\bm{x}} \bm{x}_0+\bm{S}_{\bm{u}} \bm{u}$ at trajectory level, where $\bm{S}_x$ and $\bm{S}_u$ are composed of $\bm{A}_t$ and $\bm{B}_t$, see \cite{Calinon_IET_2023} for details.

Similarly as in DMP, where basis functions are used as movement primitives to represent the non-linear forcing term, basis functions can be used to represent the corresponding control commands. By leveraging a least squares formulation of recursive LQR, we showed in \cite{Calinon_IET_2023} that the control commands can be expressed as $\bm{u} = -\bm{F} \tilde{\bm{x}}_0 = -\bm{\Psi} \bm{W} \tilde{\bm{x}}_0$ where $\bm{\Psi}$ are the basis functions stored in matrix form, $\bm{W}$ is the weight matrix, and $\tilde{\bm{x}}_0 $ is the augmented initial state for transferring the original LQT problem into an LQR problem with control primitives:
\begin{equation}\label{eq:batch_formulation_lqr}
    \begin{aligned}
        \min_{\bm{W}} 
        & = \Tilde{\bm{x}}^{\top} \Tilde{\bm{Q}} \Tilde{\bm{x}} + (\bm{\Psi} \bm{W} \tilde{\bm{x}}_0)^{\top} \bm{R} (\bm{\Psi} \bm{W} \tilde{\bm{x}}_0), \\
        &\textrm{s.t.} \quad  \tilde{\bm{x}}=(\tilde{\bm{S}}_{\bm{x}} - \tilde{\bm{S}}_{\bm{u}} \bm{\Psi} \bm{W})\tilde{\bm{x}}_0, \\
    \end{aligned}
\end{equation}
where $\Tilde{\bm{x}}$ is the concatenation of augmented state $\Tilde{\bm{x}}_t = [\bm{x}^\trsp_t, 1]^{\top}$, $\Tilde{\bm{Q}}$ is the concatenation of augmented weight matrices computed as
$$
\tilde{\bm{Q}}_t=\left[\begin{array}{cc}
\bm{I} & \mathbf{0} \\
-\bm{\mu}_t^{\top} & 1
\end{array}\right]\left[\begin{array}{cc}
\bm{Q}_t & \mathbf{0} \\
\mathbf{0} & 1
\end{array}\right]\left[\begin{array}{cc}
\bm{I} & -\bm{\mu}_t \\
\mathbf{0} & 1
\end{array}\right],
$$
$\tilde{\bm{S}}_x$ and $\tilde{\bm{S}}_u$ are  state transfer matrices composed of $\tilde{\bm{A}}_t = \left[\begin{array}{cc} \bm{A}_t & \mathbf{0} \\
\mathbf{0} & 1 \end{array}\right]$ and $\tilde{\bm{B}}_t = \begin{bmatrix} \bm{B}_t \\ 0 \end{bmatrix}$ that define the augmented system dynamics $\tilde{\bm{x}}_{t+1}=\bm{\tilde{A}}_t \tilde{\bm{x}}_t+\bm{\tilde{B}}_t \bm{u}_t$.

Solving \eqref{eq:batch_formulation_lqr} results in the optimal weight matrix estimations:
\begin{equation}
    \hat{\bm{w}} = (\bm{\Psi}^{\top} \bm{\tilde{S}_{u}}^{\top} \bm{\tilde{Q}} \bm{\tilde{S}_{u}} \bm{\Psi} + \bm{\Psi}^{\top} \bm{R} \bm{\Psi})^{-1} \bm{\Psi}^{\top} \bm{\tilde{S}_{u}}^{\top} \bm{\tilde{Q}} \bm{\tilde{S}_x},
\end{equation}

The above solution can be used as a recursive formulation to generate a feedback controller that can handle external perturbation in real-time execution, providing a \textcolor{blue}{feedback} controller
$\bm{u}_t = -\tilde{\bm{K}}_t \tilde{\bm{x}}_t$ where $\tilde{\bm{K}}_t$ is the feedback gain matrix at time step $t$. We can therefore recursively calculate the feedback gains of our feedback controller LQT-CP with
\begin{equation}
    \begin{aligned}
        \bm{\tilde{K}}_t &= \bm{\Psi}_t \hat{\bm{W}} \bm{P}_t, \\
        \bm{P}_t &= \bm{P}_{t-1} (\bm{\tilde{A}}_{t-1} - \bm{\tilde{B}}_{t-1} \bm{\tilde{K}}_{t-1})^{-1},
    \end{aligned}
\end{equation} 
where $\tilde{\bm{K}}_0 = \bm{\Psi}_0\bm{W}$, $\bm{P}_0 = \bm{I}$, see \cite{Calinon_IET_2023} for details. 

LQT-CP provides additional flexibility in the design of the precision matrix $\bm{Q}$ to encode the importance of tracking different profiles of the demonstrated trajectory as well as capturing their correlation constraints. Therefore, we can design $\bm{Q}$ to track the acceleration profiles and goal position of the demonstrated trajectory to generate an optimal control formulation of DMP. \textcolor{blue}{With a single demonstration, the precision matrix Q can be arbitrarily high as long as the relative importance of those two terms is captured, similar to the specification of stiffness and damping gains in classical DMP. While multiple demonstrations are available, one can also set the precision $\bm{Q}$ as the inverse of the observed covariance so that the system reproduces the task with the same (co)variations as in the set of demonstrations \cite{Calinon14ICRA, zhang2022learning}. We further incorporate via-point specifications by simply indicating the via-points in the reference trajectory $\bm{\mu}$ and assigning the same importance to via-point and goal reaching in the weight matrix $\bm{Q}$. This simple extension allows us to address contact-rich manipulation sub-tasks while concurrently extending LQT-CP to handle long-horizon manipulation tasks. We illustrate this modification with the example of pulling a cube using a hook in Section \ref{subsection:lqt_cp}.}


\subsection{Logic-LfD}
In this section, we introduce Logic-LfD for Task and Motion Planning (TAMP).
The primary enhancement involves augmenting the classical demonstration representation, typically consisting of geometrical state sequences (e.g., position trajectories), with logical state and action abstractions derived from PDDL, and more importantly integrating TAMP solver with the proposed LQT-CP.

Consider a long-horizon manipulation task with $N$ sub-tasks to be solved sequentially. The demonstration denoted as $\mathcal{P} = \{\{\mathcal{L}_{i}, \bm{a}_{i}\}_{i=1,\dots,N}, \{\bm{\tau}_{k}\}_{k=1,\dots, K}\}$, encompasses both task- and motion-level trajectories. Here, $\{\mathcal{L}_{i}\}_{i=1,\dots,N}$ represents the logical state sequence from the initial to the goal scene graph, akin to a task-level trajectory $\mathcal{L}$. $\bm{a}_{i}$ denotes the abstracted action, defining the transition from $\mathcal{L}_{i}$ to $\mathcal{L}_{i+1}$, and specifies which robot arm(s) executes the motion-level trajectory for manipulating which particular object(s). \textcolor{blue}{$\bm{\tau}_{k}$ represents the motion-level trajectory demonstrated for training the corresponding LQT-CP for the execution of the corresponding action primitive $\bm{a}_{k}$}. In contrast to the straightforward propagation of geometrical states through Euclidean or unit quaternion operations ($\oplus, \ominus$), the logical state propagation is confined to a feasible action set $\mathcal{A}=\{a_k\}_{k=1,\dots, K}$ for the target task. Consequently, the corresponding demonstration for Logic-LfD should include not only the logical and geometrical state sequences but also the corresponding action sequence $\{\bm{a}_{i}\}_{i=1,\dots,N}$ indicating the feasible propagation of logical states. 

Given the demonstration $\mathcal{P}$, we offline train the feedback gains of a set of LQT-CPs for each action primitive in $\mathcal{A}$ with the corresponding motion-level demonstrations $\{\bm{\tau}_{k}\}_{k=1,\dots, K}$. Replacing $\bm{\tau}_{k}$ in $\mathcal{P}$ with the corresponding LQT-CP$_k$, by exploring the generalization ability of LQT-CP \textcolor{blue}{at the motion level, we can} generate an initial plan $\mathcal{P}_{I}$ which can handle new long-horizon manipulation tasks that only require the generalization of geometrical states. However, this limited generalization ability cannot address new tasks that require generalization in both task and motion levels.

Therefore, we further propose a new integration of TAMP solver with the proposed LQT-CP method. This integration results in Logic-LfD that quickly generalizes the demonstration $\mathcal{P}$ to solve new long-horizon tasks \textcolor{blue}{which} share the same task goal state $\bm{G}$ but have different initial states than the template task, thus requiring generalization in both task and motion levels.

The overview of the whole framework is depicted in Figure \ref{fig:framework}, shown with a four-block stacking scenario. Supposing the new task starts with a new logical state $\mathcal{L}_{0}^{\prime}$, Logic-LfD efficiently solves this new task with a two-step planning process and only needs to solve a partial TAMP problem. The first step involves finding a feasible task-and-motion trajectory $\mathcal{P}^{\text{new}}$ that transitions $\mathcal{L}_{0}^{\prime}$ to its closest logical state $\mathcal{L}_{1} \in \mathcal{L}$ in the demonstration $\mathcal{P}$, then concatenating $\mathcal{P}^{\text{new}}$ with the corresponding segmentation of the initial plan $\bm{\mathcal{L}_{1}} \rightarrow  \bm{\mathcal{L}_{2}} \rightarrow \bm{G}$ to establish a feasible task plan transitions $\mathcal{L}_{0}^{\prime}$ to $\bm{G}$ within the smallest time cost. Secondly, Logic-LfD generates the feasible motion-level trajectories by exploring the generalization ability of the proposed LQT-CP. \textcolor{blue}{$\mathcal{P}^{\text{new}}$, the state closest to the new initial state $\mathcal{L}_{0}^{\prime}$} is indirectly determined by reformulating the demonstrated task-level trajectory $\mathcal{L} = \{\mathcal{L}_{i}\}_{i=1,\dots,N}$ as multiple goals $\mathcal{G}$ \textcolor{blue}{(referring to the \textit{MultiGoalsSpecification} function in the Algorithm \ref{alg:reactive_tamp})} and running the \textit{PDDLStream} \cite{garrett2020pddlstream} algorithm to reach these goals concurrently, with the one reached \textcolor{blue}{within the minimum time} considered as the target closest state. It is worth noting that our Logic-LfD framework treats the TAMP solver as a black-box tool, allowing for the integration of any TAMP solver into the framework, demonstrating its generality. 
\subsection{Reactive TAMP with Logic-LfD}
Logic-LfD ensures fast discovery of a feasible task and motion plan. Since the closest state is usually reached much earlier than the goal state, Logic-LfD often resolves only a partial TAMP planning process while reacting to task-level disturbances. Considering that the planning time of a TAMP solver tends to increase significantly with the length of the final action skeleton, solving only a partial TAMP problem allows Logic-LfD to perform notably faster than the original TAMP solver. Therefore, we further extend the Logic-LfD into a close-loop TAMP framework (Algorithm \textcolor{red}{\ref{alg:reactive_tamp}}) which promptly reacts to task- and motion-level disturbances while executing the nominal plan in dynamic environments.

It assumes the task goal $G$, demonstration $\mathcal{P}$, and an initial plan $\mathcal{P}_{I}$ are given. In the initialization process, multi-goal specifications $\mathcal{G}$ are generated based on the demonstration and the task goal with the designed \textit{MultiGoalsSpecification} function in Line 4. This function filters out the static facts in $\mathcal{L}$ and keeps the sequence of fluents in $\mathcal{P}$. The designed \textit{SceneGraph} function generates the facts about the current environment based on captured information about objects and robots in the environment and outputs the filtered key facts. 

\begin{algorithm}[t]
    \caption{Reactive TAMP with Logic-LfD}
    \label{alg:reactive_tamp}
    \textbf{Given:} Environment $env$, Demonstration $\mathcal{P}$, Task Goal $\bm{G}$, Initial Plan $\mathcal{P}_{I}$ \\
    \textbf{Output:} next action $\bm{a}$ and motion $\bm{\tau}$ \\
    \textbf{Initialization:} \\
        $\mathcal{G} = \textit{MultiGoalsSpecification}(\mathcal{P}, \bm{G})$ \\
        $\mathcal{S}_{C} = \textit{SceneGraph}(env)$ \\
        $\mathcal{P}_{C} \gets \mathcal{P}_{I} $ \\
    \While{$\bm{G} \not \in \mathcal{S}_{C}$}{
        $flag, id = \textit{LogicIn}(\mathcal{S}_{C}, \mathcal{G})$ \\
        \eIf{flag is True}{
          $\bm{a} = \mathcal{P}_{I}[id]$ \\
        }{
          $\mathcal{P}^{\text{new}} = \textit{PDDLStream}(\mathcal{S}_{C}, \mathcal{G})$\\
        $\bm{a} = \mathcal{P}^{\text{new}} \left[0\right]$ \\
        }
        $\bm{\tau} = LQT\_CP(\bm{a})$ \\
        $\mathcal{S}_{C} = \textit{SceneGraph}(env)$ \\
    }
\end{algorithm}

To solve the target task in \textcolor{blue}{a} closed loop, we continuously estimate the \textcolor{blue}{current} logical states $\mathcal{S}_{C}$ after executing each action $\bm{a}$ in the current plan $\mathcal{P}_{C}$. If the target \textcolor{blue}{goal} $\bm{G}$ is not a subset of $\mathcal{S}_{C}$, indicating the task is not achieved yet. We then invoke the Logic-LfD to swiftly generate a new plan. With the \textit{LogicIn} function in Line 8, we check if \textcolor{blue}{the} current state $\mathcal{S}_{C}$ is a subset of the multi-goal specifications $\mathcal{G}$. If it is, we also generate its corresponding index \textit{id} in the sequence. Then, the action $\mathcal{P}_{I}[id]$ in the initial plan will be a feasible action that transitions \textcolor{blue}{the} current state $\mathcal{S}_{C}$ to the goal state $\bm{G}$. If $\mathcal{S}_{C}$ is not a subset of $\mathcal{G}$, it indicates a severe task-level disturbance in the environment, and \textit{PDDLStream} with multi-goal specification $\mathcal{G}$ is applied to find a new feasible plan $\mathcal{P}^{\text{new}}$ that converges to the demonstration $\mathcal{P}$. The first action in $\mathcal{P}^{\text{new}}$ is executed as the feasible action for the current state. The action $\bm{a}$ not only shows which action should be executed, but also indicates which robot arm(s) should execute that action for manipulating which object(s). The LQT-CP is then applied in an object-centric manner to generalize the reference trajectory for generating the corresponding motion-level trajectory $\bm{\tau}^{\text{new}}$ to execute action $\bm{a}$.

\section{Experiments}\label{sec:experiments}
\begin{figure}[!htb]
    \centering
    \begin{subfigure}{0.15\textwidth}
        \centering
        \includegraphics[width=1\linewidth]{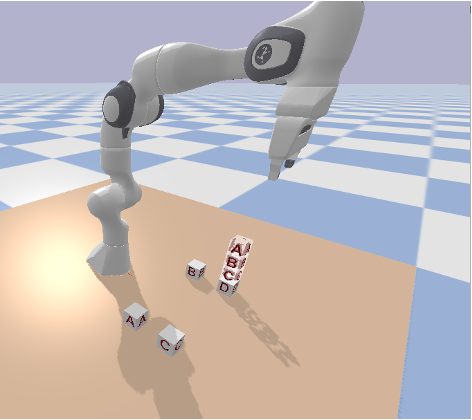}
        \caption{B1}
        \label{fig:exp1}
    \end{subfigure}%
    \hspace{0.02\linewidth}
    \begin{subfigure}{0.15\textwidth}
        \centering
        \includegraphics[width=1\linewidth]{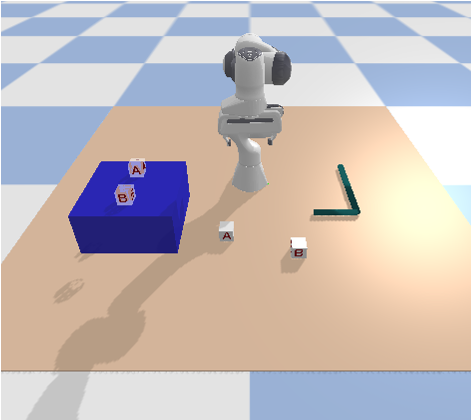}
        \caption{B2}
        \label{fig:exp2}
    \end{subfigure}%
    \hspace{0.02\linewidth}
    \begin{subfigure}{0.15\textwidth}
        \centering
        \includegraphics[width=1\linewidth]{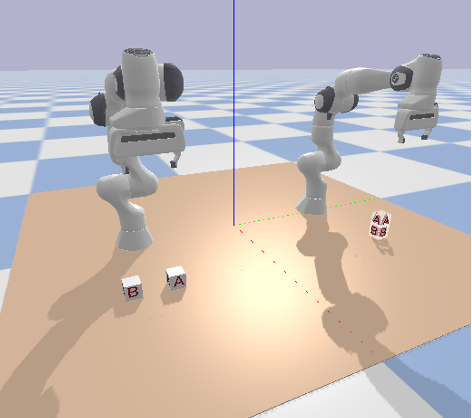}
        \caption{B3}
        \label{fig:exp3}
    \end{subfigure}
    \caption{Experimental setups for the three benchmarks. \textit{B1}: Tower Construction, \textit{B2}: Workspace Reach, \textit{B3}: Dual-arm Box Transportation. Each sub-figure illustrates the demonstrated task, with the initial and task goal \textcolor{blue}{states} depicted in grey and white color, respectively.}
    \label{fig:experimental_setups}
\end{figure}

In this section, we present a comparative analysis between Logic-LfD and several baselines across three long-horizon manipulation planning problems. Figure~\ref{fig:experimental_setups} shows the experiment setups and the demonstrated tasks for the benchmarks. For each benchmark, we demonstrated the robot with an initial task and motion plan based on available action sets. Subsequently, we introduced distinct initial setups or disturbances to assess the performance of Logic-LfD against the selected baselines. 

\subsection{Benchmarks}\label{subsection:benchmarks}
\subsubsection{Tower Construction (\textbf{B1})}
This experiment involves controlling a robotic arm to manipulate a set of blocks using actions from the set $\mathcal{A} = \{pick, place, stack, unstack\}$ to arrange them in a specific order. Using the example of constructing a four-block tower, the template task begins with an initial state where all four blocks are placed on the table (i.e., without stacked blocks). The task goal state is achieved when the blocks are stacked in the sequence \textit{A on B on C on D}. During the interaction between the robot and \textcolor{blue}{the} blocks, both task- and motion-level constraints should be considered. For instance, the action \textit{pick} for block \textit{A} is only permissible if it is \textit{clear} (no blocks on it) and if a \textit{collision-free} trajectory exists. 

\subsubsection{Workspace Reach (\textbf{B2})}
As an extension of the Tower Construction benchmark, this scenario involves the robot arm manipulating blocks with tools to address more complex long-horizon manipulation tasks. Blocks may be positioned beyond the reach of the robot arm. The template task begins with an initial state where a few blocks and one hook are placed on the table within the robot arm's workspace. It ends with the task goal state, where all blocks are placed on the shelf on the table. Actions \textit{stack} and \textit{unstack} are excluded to simplify the definition of the \textit{pull} action which describes instances where the hook is grasped by the robot arm to extend its workspace and pull the block within its original workspace for picking or placing. 

\subsubsection{Dual-arm Block Transportation (\textbf{B3})}
This benchmark further extends the Tower Construction experiment to validate the proposed algorithm in more complex multi-arm scenarios. The feasible action set $\mathcal{A} = \{pick, place, stack, unstack\}$ remains the same, but with augmented parameters introducing two \textit{arms}. This complexity leads to more instantiations of abstracted actions and results in a more intricate long-horizon manipulation planning task. The template task begins with an initial state where all blocks are on the left table without any blocks on top and ends with the task goal state where blocks are expected to be stacked in a given sequence as in B1.

\subsection{Disturbances}
To comprehensively evaluate the generalization capability and reactivity of Logic-LfD across all benchmarks, we apply variants on initial states or disturbances at any states at different levels.

\begin{itemize}
    \item [L1] \textit{Motion-level Variant/Disturbance:} Blocks are subjected to disturbances in different positions, while logical states remain consistent with the original or expected conditions;
    \item [L2] \textit{Slight Task-level Variant/Disturbance:} Blocks undergo disturbances, resulting in a different logical state. This logical state aligns with that seen in the demonstration for the template task;
    \item [L3] \textit{Severe Task-level Variant/Disturbance:} Blocks are disturbed to a novel and previously unseen logical state in the demonstration;
    \item [L4] \textit{Extreme Task-level Variant/Disturbance:} A new block is introduced into the scene, \textcolor{blue}{thus significantly influencing} the execution of the initial plan to achieve the target goal.
\end{itemize}

\subsection{Comparison between LQT-CP and DMP}\label{subsection:lqt_cp}
\begin{figure}[thb]
    \centering
    \begin{subfigure}{0.5\textwidth}
        \includegraphics[trim=1.5cm 3.8cm 1.5cm 3.8cm, clip, width=\linewidth]{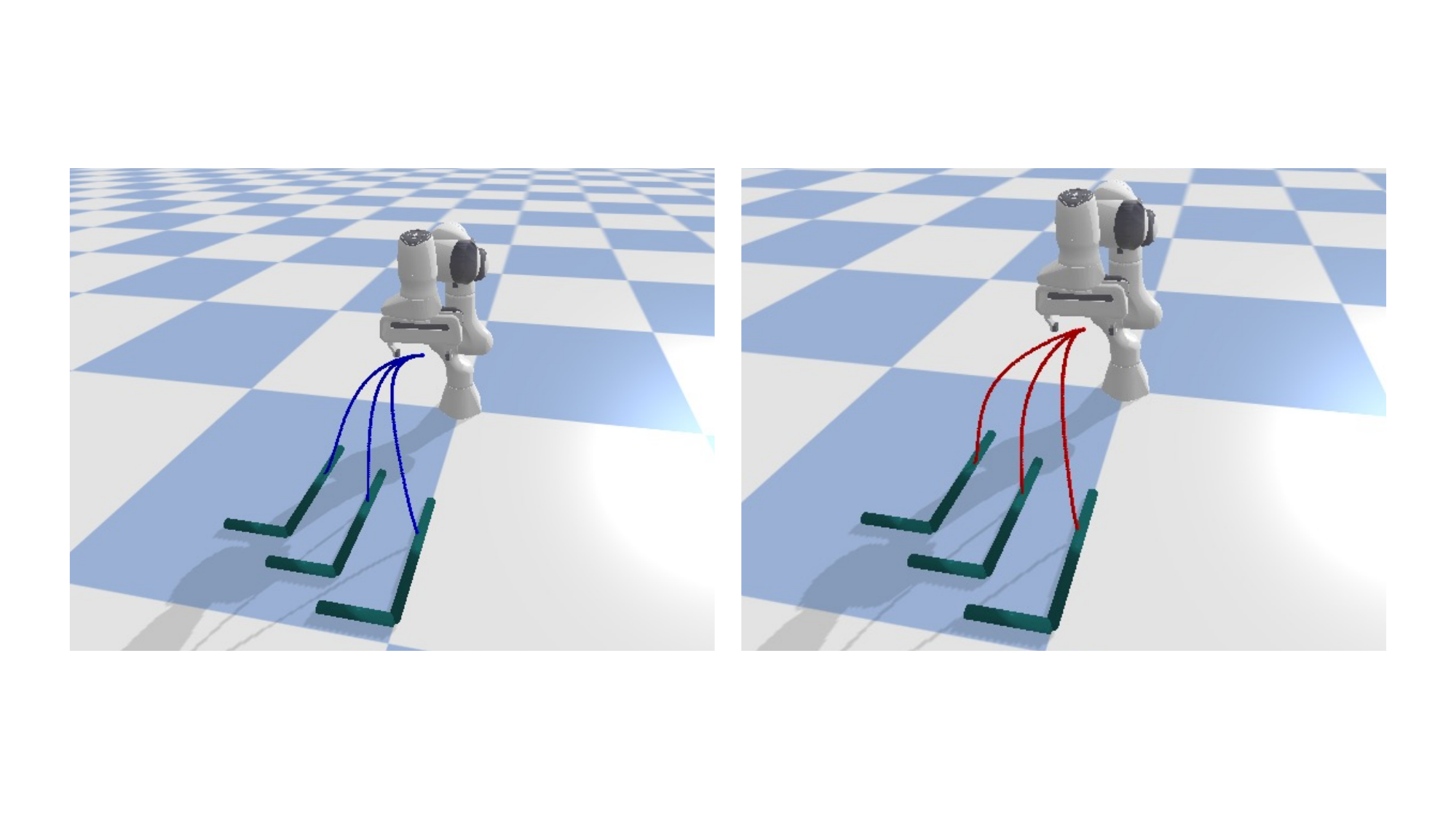}
        \label{fig:pick_hook}
    \end{subfigure}
    \begin{subfigure}{0.5\textwidth}
        \includegraphics[trim=1.5cm 3.8cm 1.5cm 3.8cm, clip, width=\linewidth]{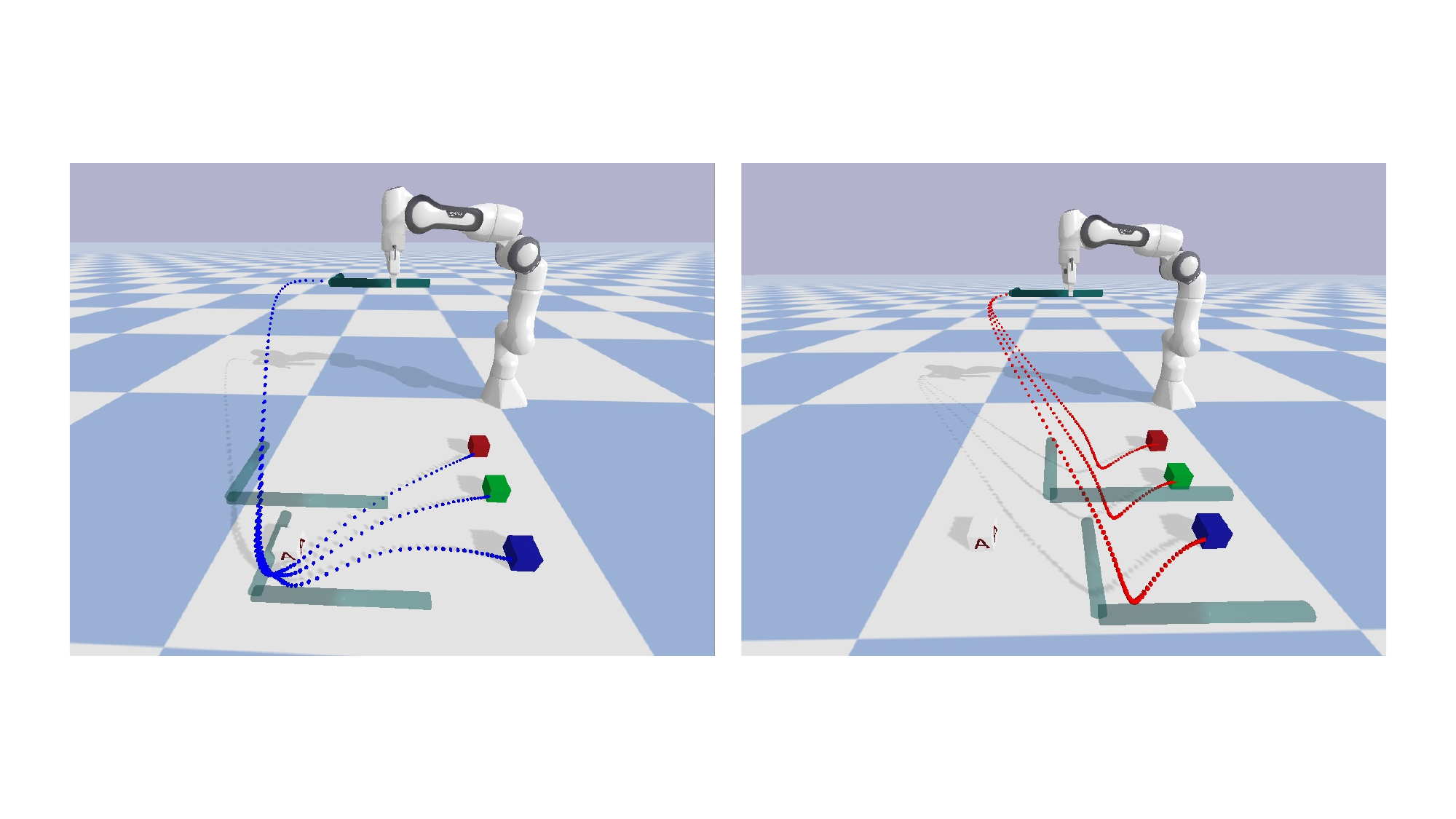}
        \label{fig:pull_hook}
    \end{subfigure}
    \vspace{-0.5cm}
    \caption{Comparison between LQT-CP and standard DMP for two sub-tasks in the Workspace Reach benchmark (B2). \textit{Top:} pick the hook. \textit{Bottom:} pull cube A with the hook. The trajectories generated from LQT-CP are illustrated with blue lines in the left figures, \textcolor{blue}{and} the ones for DMP are \textcolor{blue}{shown} with red lines \textcolor{blue}{on} the right. In the pulling task, the hook is expected to pass through two crucial via-points, the top and left-down corners of the cube A, for successfully accomplishing the task. In this figure, only the via-points for pulling cube A to the blue goal are shown with transparent hooks in the \textit{Bottom} figures.}
    \label{fig:lqt_vs_dmp}. 
\end{figure}

In this section, we demonstrate the flexible motion modulation capabilities of LQT-CP, addressing two crucial sub-tasks in B2. The experimental results are depicted in Figure~\ref{fig:lqt_vs_dmp}. In the initial picking sub-task, the robot arm is tasked with picking up a hook potentially placed at three distinct positions. Both LQT-CP and DMP exhibit comparable performance, effectively generalizing the reference trajectory to new goal positions, as showcased in Figure~\ref{fig:lqt_vs_dmp}-\textit{Top}.

In the subsequent pulling sub-task, the robot arm is required to pull cube A using the hook to three different goal positions represented by red, green, and blue cubes in Figure~\ref{fig:lqt_vs_dmp}-\textit{Bottom}. This task demands not only the generalization of the reference pulling trajectory to new goal positions but also the precise control of the hook passing through two crucial via-points (the top and the left-down corners of cube A) for successful pulling. Consequently, it necessitates more flexible motion modulation concerning via-points tracking and generalization. Figure~\ref{fig:lqt_vs_dmp}-\textit{Bottom} illustrates the actual position of the hook when expected to be at the via-points. Notably, LQT-CP adeptly generalizes the reference trajectory and via-points, successfully pulling cube A to the goal positions, while standard DMP encounters challenges, which highlights the flexibility of the proposed LQT-CP formulation of DMP for motion modulation.

\subsection{Generalization Ability of Logic-LfD}\label{subsection:logic_dmp_generalization}  
In this section, we assess the generalization ability of Logic-LfD in comparison to two baseline methods: \textit{Linear} and \textit{PDDLStream}. Here, \textit{Linear} refers to the linear execution of a fixed sequence of DMPs without adapting the action sequence when addressing new tasks. For a comprehensive evaluation, each benchmark comprises tasks that share the same task goal but vary in their initial states from the template task. For B1, we solved a four-block tower construction problem. \textcolor{blue}{For B2 and B3, tasks were simplified by utilizing only two blocks.} Moreover, LQT-CP serves as the motion modulation method for all baselines to ensure a fair comparison. We analyzed the overall success rate and computation time of the three methods in solving 100 tasks with random initial states for each benchmark across various levels (L1-L4). The summarized experimental results are presented in Table~\textcolor{red}{\ref{tab:generalization}}.

In summary, both Logic-LfD and PDDLStream successfully solved all the validation tasks, while the linear execution of DMPs achieves success in only approximately $25\%$ of the tasks, primarily those featuring motion-level variants in the initial states. The results show that Logic-LfD can enhance the generalization ability beyond linear execution. Moreover, Logic-LfD exhibits faster planning capabilities than PDDLStream across all benchmarks, with a $30\%$ to $40\%$ improvement in B2 and B3, and a remarkable $70\%$ improvement in B1. The substantial acceleration observed in  B1 can be attributed to the inherently longer action sequence in the task, with which the planning time of PDDLStream increases significantly. Since Logic-LfD only needs to partially address the TAMP problem with a shorter action skeleton, it notably decreases the whole planning time, showing a significant acceleration in the experimental results. This reinforces the efficacy of Logic-LfD in handling long-horizon manipulation tasks and motivates us to formulate the reactive TAMP framework in Algorithm~\textcolor{red}{\ref{alg:reactive_tamp}} by employing Logic-LfD in a closed-loop manner.

\subsection{Reactivity of TAMP with Logic-LfD}
In this section, we conduct a comparative analysis of the reactivity exhibited by the proposed closed-loop Logic-LfD framework and two baseline approaches: \textit{Linear} and \textit{RLDS}. The \textit{Linear} baseline involves the linear execution of a predetermined sequence, without closed-loop feedback. \textit{RLDS} is a reproduction of the algorithm outlined in \cite{Paxton_Ratliff_Eppner_Fox_2019}, with the incorporation of the proposed LQT-CP for motion modulation. The experimental setup aligns with that detailed in Section~\ref{subsection:logic_dmp_generalization}, ensuring consistency across benchmarks. We assess the performance of each method in handling disturbances at different levels (L1-L4) and report the average execution times based on 10 trials for each disturbed task in Table~\ref{tab:reactive}.
\begin{table*}[thb!]
    \centering
    \begin{tabularx}{\textwidth}{|X|X|X|X|X|X|X|X|X|X|X|X|}
        \hline
        \multicolumn{3}{|c|}{} & \multicolumn{3}{c|}{\textbf{Linear / Sequential}} &  \multicolumn{3}{c|}{\textbf{PDDLStream}} & \multicolumn{3}{c|}{\textbf{Logic-LfD}} \\
        \cline{4-12}
        \multicolumn{3}{|c|}{} & \textbf{B1} & \textbf{B2} & \textbf{B3} & \textbf{B1} & \textbf{B2} & \textbf{B3} & \textbf{B1} & \textbf{B2} & \textbf{B3}\\
        \hline
        \multicolumn{3}{|c|}{\textbf{Success Rate}} & $27\%$ & $31\%$ & $24\%$ & $100\%$ & $100\%$ & $100\%$ & \textbf{100\%} & \textbf{100\%} & \textbf{100\%} \\
        \hline
        \multicolumn{3}{|c|}{\textbf{Time [s]}} & N/A & N/A & N/A & $0.53\pm0.10$ & $0.53\pm0.23$ & $0.23\pm0.09$ & \textbf{0.16}$\bm{\pm}$\textbf{0.07} & \textbf{0.37}$\bm{\pm}$\textbf{0.27} & $\textbf{0.14}\bm{\pm}\textbf{0.06}$ \\
        \hline
    \end{tabularx}
    \caption{Generalization ability comparison of Logic-LfD with baselines when solving tasks with various initial states in benchmarks. 'N/A' means no feasible values for the corresponding elements. Logic-LfD demonstrates superior generalization ability, achieving a higher success rate than the \textit{Linear} execution of DMPs, and incurs significantly smaller computation time compared to PDDLStream.}
    \label{tab:generalization}
\end{table*}

\begin{table*}[thb!]
    \centering
    \begin{tabularx}{\textwidth}{|X|X|X|X|X|X|X|X|X|X|X|X|}
        \hline
        \multicolumn{3}{|c|}{} & \multicolumn{3}{c|}{\textbf{Linear / Sequential}} & \multicolumn{3}{c|}{\textbf{RLDS}} & \multicolumn{3}{c|}{\textbf{Logic-LfD}} \\
        \cline{4-12}
        \multicolumn{3}{|c|}{} & \textbf{B1} & \textbf{B2} & \textbf{B3} & \textbf{B1} & \textbf{B2} & \textbf{B3} & \textbf{B1} & \textbf{B2} & \textbf{B3} \\
        \hline
        \multicolumn{3}{|c|}{\textbf{L1}} & 13.23$\pm$0.58 & $12.50\pm0.13$ & $24.63\pm0.05$ & 13.70$\pm$1.23 & 12.59$\pm$0.19 & 24.64$\pm$0.04 & \textbf{13.23}$\bm{\pm}$\textbf{0.35} & \textbf{12.50}$\bm{\pm}$\textbf{0.13} & \textbf{24.62}$\bm{\pm}$\textbf{0.03}\\
        \hline
        \multicolumn{3}{|c|}{\textbf{L2}} & N/A & N/A & N/A & 18.07$\pm$0.95 & 19.08$\pm$0.01 & 30.83$\pm$0.04 & \textbf{17.62}$\bm{\pm}$\textbf{0.67} & \textbf{19.08}$\bm{\pm}$\textbf{0.01} & \textbf{30.79}$\bm{\pm}$\textbf{0.02}\\
        \hline
        \multicolumn{3}{|c|}{\textbf{L3}} & N/A & N/A & N/A & N/A & N/A & N/A & \textbf{23.05}$\bm{\pm}$\textbf{0.49} & \textbf{26.03}$\bm{\pm}$\textbf{0.78} & \textbf{30.23}$\bm{\pm}$\textbf{0.08}\\
        \hline
        \multicolumn{3}{|c|}{\textbf{L4}} & N/A & N/A & N/A & N/A & N/A & N/A & \textbf{19.00}$\bm{\pm}$\textbf{1.29} & \textbf{12.60}$\bm{\pm}$\textbf{0.15} & \textbf{31.57}$\bm{\pm}$\textbf{0.04}\\
        \hline
    \end{tabularx}
    \caption{Comparison of average execution time $[s]$ between Logic-LfD and baselines in three benchmarks. 'N/A' denotes no successful trials. Closed-loop Logic-LfD shows superior reactivity to various disturbances in all benchmarks. Please refer to the supplementary video through the project webpage for the illustration of simulation experiments for each element in the table.}
    \label{tab:reactive}
\end{table*}
\begin{figure*}[thb]
    \centering
    \includegraphics[trim=0.92cm 7.4cm 0.92cm 7.4cm, clip, width=\linewidth]{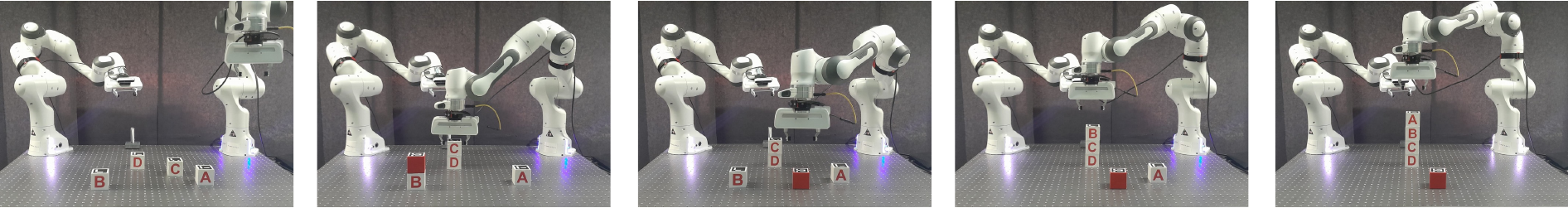}
    \caption{Closed-loop Logic-LfD under extreme task-level disturbance (placing the red block on cube B after stacking cube C on D) in a real-world four-block stacking task. Logic-LfD reacts to the disturbance by unstacking the red block from cube B and \textcolor{blue}{placing} it on the table, then continuing the original plan for achieving the goal state, as shown in the last figure.}
    \label{fig:real_world}
\end{figure*}

Our analysis reveals that the proposed closed-loop Logic-LfD and RLDS demonstrate comparable reactivity in the presence of motion-level and slight task-level disturbances. Specifically, under L1 and L2 disturbances, both closed-loop Logic-LfD and RLDS effectively respond to the disturbances and achieve task goals within comparable execution times. However, RLDS relies on backward checking of the planned action sequence without plan modification, leading to challenges in handling disturbances at L3 and L4 across all benchmarks. In contrast, Logic-LfD exhibits the ability to generate a new feasible action plan when faced with those disturbances, enabling the TAMP approach to accomplish tasks even under severe or extreme disturbances. Notably, Logic-LfD proves to be faster in deriving the new action plan compared to the straightforward application of TAMP solvers, as demonstrated in the comparison results between PDDLStream and Logic-LfD in Section~\ref{subsection:logic_dmp_generalization}, showing the significant reactivity inherent to the proposed closed-loop Logic-LfD approach.

\subsection{Real-world Experiments}
We assessed the reactivity of the proposed Logic-LfD in closed-loop scenarios for handling various task-level disturbances within the four-block Tower Construction scenario, as shown in Figure~\ref{fig:real_world}. All experiments were conducted using a 7-axis Franka Emika robot arm equipped with a RealSense D435 camera for object localization. The entire planning algorithm demonstrates the capability to respond to severe or extreme task-level disturbances at the frequency of 1Hz, providing corresponding actions and generalized motion sequences for rearranging the objects into the desired goal configuration. Additional demonstrations of the reactive behaviors under various task-level disturbances can be found in the supplementary video of the project webpage.

\section{Conclusion}
In summary, we introduced Logic-LfD, an Learning from Demonstration (LfD) approach tailored for long-horizon manipulation tasks in dynamic environments. We leveraged an optimal control formulation of Dynamical Movement Primitive (DMP), Linear Quadratic Tracking with Control Primitives (LQT-CP), which naturally extends DMP to incorporate via-point specifications, proving beneficial for handling contact-rich manipulation sub-tasks. We demonstrated that Logic-LfD shows superior generalization ability than DMP and superior efficiency than TAMP solver in handling task variants during long-horizon planning. Consequently, we further extended this framework into a reactive TAMP system to quickly react to disturbances in dynamic environments. We validated the proposed methods through various simulation and real-world experiments, affirming the performances in term of skill transfer, generalization ability, and reactiveness to disturbances in long-horizon manipulation tasks.  

In this paper, we integrated DMP with TAMP solver because of its exceptional extrapolation generalization ability with one single demonstration. Notably, this integration is not confined to DMP alone. It can be extended to other LfD algorithms. An interesting avenue involves further extending Logic-LfD by integrating with diffusion models \cite{chi2023diffusion} to tackle more complex long-horizon contact-rich manipulation tasks. Moreover, we aim to extend the proposed method to other real-world scenarios characterized by partially observable environments, which necessitate rapid replanning based on new observations.

One drawback of Logic-LfD is its potential to find slightly longer solutions when generalizing to new tasks compared to directly applying \textcolor{blue}{a} TAMP solver. This is attributed to the consideration of goals in multi-goal specifications with the same priority. When multiple feasible solutions exist, Logic-LfD selects the one reached in the fastest way, but it doesn't guarantee that the chosen goal is the closest to the task goal state among all feasible solutions. Future work should explore strategies for assigning priorities to goals within the multi-goal specifications and the integration with optimization-based TAMP solvers \cite{toussaint2015logic, xue2023d} to solve long-horizon manipulation tasks where the optimality of metrics (e.g. time and energy efficiency) \textcolor{blue}{is} important. 

\bibliography{IEEEabrv,main}
\bibliographystyle{IEEEtran}
\newpage
\end{document}